\documentclass{edm_article}

\usepackage{amsmath}
\usepackage{graphicx}
\usepackage{url}
\usepackage{multirow}
\usepackage{tabularx}
\usepackage{rotating}
\usepackage{booktabs} 
\usepackage{xargs} 
\usepackage[colorinlistoftodos,prependcaption,textsize=tiny]{todonotes}
\newcommandx{\annanotes}[2][1=]{\todo[inline,linecolor=green,backgroundcolor=green!5,bordercolor=green,#1]{#2}}
\usepackage{comment}

\begin{document}

\title{Getting too personal(ized): The importance of feature choice in online adaptive algorithms}  
\date{}

\numberofauthors{6}
\author{
\alignauthor
ZhaoBin Li\\
    \affaddr{Carleton College}\\
    \email{liz2@carleton.edu}\\ 
\alignauthor
Luna Yee\\
    \affaddr{Carleton College}\\
    \email{yeec@carleton.edu}
\alignauthor
Nathaniel Sauerberg\\
    \affaddr{Carleton College}\\
    \email{sauerbergn@carleton.edu}
\and
\alignauthor
Irene Sakson\\
    \affaddr{Carleton College}\\
    \email{saksoni@carleton.edu}
\alignauthor
Joseph Jay Williams\\
    \affaddr{University of Toronto}\\
    \email{williams@cs.toronto.edu}
\alignauthor
Anna N. Rafferty\\
    \affaddr{Carleton College}\\
  \email{arafferty@carleton.edu}
}
\maketitle

\begin{abstract}    
Digital educational technologies offer the potential to customize students' experiences and learn what works for which students, enhancing the technology as more students interact with it. We consider whether and when attempting to discover how to personalize has a cost, such as if the adaptation to personal information can delay the adoption of policies that benefit all students. We explore these issues in the context of using multi-armed bandit (MAB) algorithms to learn a policy for what version of an educational technology to present to each student, varying the relation between student characteristics and outcomes and also whether the algorithm is aware of these characteristics. 
Through simulations, we demonstrate that the inclusion of student characteristics for personalization can be beneficial when those characteristics are needed to learn the optimal action. In other scenarios, this inclusion decreases performance of the bandit algorithm. Moreover, including unneeded student characteristics can systematically disadvantage students with less common values for these characteristics. Our simulations do however suggest that real-time personalization will be helpful in particular real-world scenarios, and we illustrate this through case studies using existing experimental results in ASSISTments~\cite{selent2016assistments}. Overall, our simulations show that adaptive personalization in educational technologies can be a double-edged sword: real-time adaptation improves student experiences in some contexts, but the slower adaptation and potentially discriminatory results mean that a more personalized model is not always beneficial. 


\end{abstract}

\section{Introduction} 

Within educational technologies, there are a myriad of ways to design instructional components such as hints or explanations. Research in education and the learning sciences provides some insight into how to design these resources (e.g.,~\cite{shute2008focus,aleven2016help}). However, there is often uncertainty about which version of a resource will be most effective in a particular context, and effectiveness may vary based on students' characteristics, such as prior knowledge or motivation. 

Randomized experiments are one way to compare multiple versions of a technology, but such experiments impose a delay between collecting required evidence and using that evidence to improve student experiences. Recently, multi-armed bandit (MAB) algorithms have been proposed to improve technologies in real time: each student is assigned to one version of the technology, and the algorithm observes the student's learning outcome~\cite{liu2014trading,williams2018enhancing}. Each subsequent student is more likely to be assigned to a version of the technology that has been more effective for previous students, as the algorithm discovers what is effective. Such algorithms maintain uncertainty as they learn, balancing exploring to learn more about what works with exploiting the observed results from previous students.
Typical MAB algorithms do not take into account student characteristics and thus can only identify which version of a technology is better for students on average, but contextual MAB algorithms can personalize which version to assign to each student, potentially increasing the number of students who are directed to versions that are most helpful for them individually~\cite{shaikh2019balancing}. 

While deploying contextual MAB algorithms could improve student experiences, it raises two potential issues. First, instructional designers must decide which student characteristics will be considered for personalization.
For instance, more concrete examples might be more helpful for students with lower prior knowledge, while more abstract examples could be more helpful for students with higher prior knowledge. This relationship could only be learned if the algorithm has `prior knowledge' as a feature of each student. Should the algorithm also consider which prerequisite course was taken when selecting an example, or is prior knowledge sufficient? Designers are unlikely to be certain which characteristics influence effectiveness, but the choice of characteristics will influence the performance of the algorithm. Excluding characteristics that do impact effectiveness could decrease the positive impact on students, but including extraneous characteristics that do \textit{not} impact effectiveness could also decrease this impact. 
In the latter case, the system might have to do more exploration to learn how the effectiveness of instruction differs along each extraneous characteristic, and so direct a greater number of students to less effective versions.

The second issue raised by online adaptive algorithms is whether the constantly adapting system will benefit certain groups of students more than others.
Since contextual MAB algorithms learn by observing how the consequences of their choices are related to feature values, students whose characteristics are less common may be more likely to interact with the algorithm when it has limited information about what is most effective for that type of student. This could exacerbate differences in outcomes between subgroups of students. Yet, such algorithms could also have an equalizing effect for students with less common characteristics: students have the potential to experience a version of the technology that is most appropriate for them, even when this version is not the most appropriate for a typical student.

In this paper, we use simulations to explore these issues and their consequences for student experiences in adaptive educational technologies which use MAB algorithms. We focus on three common types of models for how student characteristics are related to outcomes: a \textit{baseline} model in which student characteristics do not impact the effectiveness of different versions of the technology; a \textit{universal optimal action} model, in which student characteristics impact effectiveness but the same version is most effective for all students; and a \textit{personalized optimal action} model, in which student characteristics impact which version leads to the best outcomes for a given student. 

We show that including the potential for personalization can negatively influence student outcomes except in the \textit{personalized optimal action} model, where this information is necessary to encode the best policy. 
The more characeristics that are included for personalization, the greater the negative impact on outcomes. Additionally, including these characteristics may lead the algorithm to systematically treat students differently based on characteristics that do not influence their outcomes. When student characteristics are not uniformly distributed, including extraneous characteristics means that students in a minority group are more negatively affected than students in the majority group.
We use experimental data to show the potential benefits of personalization and add nuance to the prior simulation results by demonstrating how personalization can benefit not only students in a minority group but also all groups of students when information about the student characteristics magnifies differences between conditions.
We end by discussing the consequences of these results for integrating adaptive components into existing educational technologies.


\section{Related Work}

A wide array of work has focused on using MAB and contextual MAB algorithms for optimization, including applications in advertising and recommendations (e.g.,~\cite{li2010contextual}), crowdsourcing (e.g.,~\cite{jain2014multiarmed}), and designing experiments and clinical trials (e.g.,~\cite{villar2015multi}). Within educational technologies, MAB algorithms have been primarily used in two ways. Some work has used these algorithms to select problems that are of an appropriate difficulty level for a particular student~\cite{clement2015multi,lan2016contextual,segal2018combining}; unlike our work, these applications typically combine learned profiles about students with a second source of knowledge, such as prerequisite structure. 
We focus on a second proposed usage of MAB algorithms in education: assigning students to a particular version of a technology. For example, non-contextual MAB algorithms have been used to choose among crowdsourced explanations~\cite{williams2016axis} and to explore an extremely large range of interface designs~\cite{lomas2016interface}. Some of this work has also considered the implications of collecting experimental data via MAB algorithms on measurement and inference~\cite{liu2014trading,rafferty2019statistical}, showing systematic biases that can impair the drawing of conclusions about the conditions. Only a limited amount of work has applied contextual MAB algorithms to personalize which versions of a technology a student experiences (e.g.,~\cite{shaikh2019balancing}, but focused primarily on measurement). We build on this body of work by considering the performance implications of several common scenarios for how student characteristics, versions of an educational technology, and outcomes are related. Additionally, by specifically examining some scenarios in which student characteristics are unevenly distributed, we raise issues about personalization for minority groups of students. 

There is a great deal of theory-based literature on both standard and contextual MAB algorithms related to quantifying performance, especially in terms of asymptotically bounding growth in cumulative regret (the amount that the expected reward from choosing an optimal action outpaces reward from the actually chosen actions). The optimal worst-case bound on regret growth is logarithmic~\cite{auer2002using}.
Furthermore, the inclusion of contextual variables increases cumulative regret at least linearly; for Thompson sampling, which we use in our simulations, the regret bounds grow quadratically in the number of contextual variables~\cite{agrawal2013thompson}.  
We use simulations to consider non-asymptotic settings and focus on areas less explored theoretically, like impacts on individual groups of students and variability in performance.

In this paper, we are particularly concerned with how outcomes differ among different groups of students. One of the promises of educational technologies is to boost all students' outcomes to the level that can be achieved by individualized tutoring~\cite{corbett2001cognitive}, and online adaptive algorithms may make it easier to develop such systems. Yet, the broader machine learning community  has recently highlighted how automated systems can learn or exacerbate existing inequalities (see, e.g.,~\cite{hajian2016algorithmic} for an overview). 
Within educational data mining, there have been mixed results when the fairness of different models has been explored, and this variation has often been correlated with the diversity of the training data: \cite{hutt2019evaluating} demonstrated that a model trained on a large and diverse dataset performed similarly well for predicting on-time graduation for students in different demographic groups, while  \cite{gardner2019evaluating} found disparities across genders in predicting course dropout, often associated with gender imbalances in the training data. This raises the issue of how to best use educational data mining in ways that promote equity across students.
Within the MAB literature specifically, there has been limited discussion of fairness (e.g.,~\cite{joseph2016fairness}), although~\cite{pmlr-v75-raghavan18a} show that a particular technical definition of data diversity can lead to fairer outcomes. Like in our work,~\cite{pmlr-v75-raghavan18a} shows cases where the presence or absence of a majority group can help or harm minority group outcomes. Our work considers scenarios specific to education, demonstrating that the particular scenario in~\cite{pmlr-v75-raghavan18a} can be generalized considerably, and more precisely characterizes the circumstances in which including personal characteristics increases equity versus where doing so may lead to systematically poorer experiences for students in a minority group.



\section{Contextual MAB Algorithms}

We treat the problem of determining what version of an educational technology will be most effective for a student as a MAB problem. In such problems, a system must repeatedly choose among several actions, $a_{1},\ldots,a_{K}$. The system initially does not know which action is likely to be the most effective, but after each action choice, the system receives feedback in the form of a stochastic reward $r^{(t)}$. 

There are a variety of MAB algorithms for choosing actions. We focus on Thompson sampling~\cite{agrawal2012Analysis}, which is a regret-minimizing algorithm that exhibits logarithmic regret growth. 
Thompson sampling maintains for each action a distribution over reward values. This distribution is updated after each action choice and represents the posterior distribution over reward values given the observed data.
At each timestep, the algorithm samples from the posterior distribution over rewards for each action, and then chooses the action with the highest sampled value.
While Thompson sampling is also applicable to real-valued rewards, many educational outcomes are binary, such as whether a student completes a homework assignment or answers a question correctly. Thus, we focus on these binary rewards in this paper, using a Beta prior distribution to enable simple conjugate updates after each choice. 


In our setting in which we choose versions of an educational technology for each student, the actions are the different versions of the technology, and the reward is the student outcome. For example, imagine a student interacting with a system to do her math homework. The system might choose between two actions when the student asks for a hint: (a) show a fully worked example, versus (b) provide the first step of the problem as a hint and ask the student to identify an appropriate second step. The student outcome could be whether or not she completes the homework assignment. 

In a traditional MAB problem, the reward distribution is fixed given the action choice. However, in the situation above, the reward may be dependent on the characteristics of the student. For instance, a student who has stronger proficiency in the prerequisite skills may be more prepared to identify what to do next in the problem, while a student with weaker proficiency may not be able to identify what to do next. A contextual MAB algorithm incorporates such student characteristics as features into its action choices.

For parametric contextual MAB algorithms, the features must be predetermined, including whether interactions between features is permitted. 
We adopt a contextual Thompson sampling approach that uses regularized Bayesian logistic regression to approximate the distribution of rewards given the features~\cite{agrawal2013thompson,chapelle2011empirical}. The algorithm learns a distribution over the feature weights as coefficients using a Gaussian posterior approximation. To make each new action choice, the algorithm computes a reward value for each action by sampling each weight independently. The chosen action is the action with the highest sampled reward value. Updates may occur after each action or in batches to decrease computational costs; because the feature vectors that we consider are relatively small, we update after each action.

\section{Importance of feature choice}

When using a contextual MAB algorithm to personalize student experiences in an educational technology, the system designer must choose which student characteristics to include as variables for personalization. The designer is very unlikely to know with certainty which student features are truly relevant and will actually impact student outcomes.
One could include every possible relevant feature, knowing that while the algorithm can learn that an included feature is not relevant, it cannot learn that a non-included feature is in fact relevant.
However, asymptotic growth rates for regret are quadratic in the number of features~\cite{agrawal2013thompson}, meaning that as more features are included, the algorithm will tend to take longer to learn. Designers thus must balance the desire to include all features that influence outcomes with the knowledge that extraneous features could hurt performance.

To better understand how student outcomes are impacted by the choice of features for personalization, we systematically explore the inclusion of both relevant and non-relevant features in a contextual MAB algorithm and examine the impact on student outcomes and on the rate of assigning students to their personally optimal version of the technology. 
For these simulations, we assume that features are uncorrelated and that their values are chosen uniformly at random for each student, i.e., the probability of observing any particular combination of features is the same as observing any other combination of features.

\subsection{Methods}

\subsubsection{Representing student features}

We focus on binary student features and thus feature values implicitly group students. For example, some CS classes may have two different prerequisites, such as a discrete math course taught by the CS department or a similar one taught by the math department. Students who have taken the CS version will all have the same value for the prerequisite feature, while those who take the math one will have the other.\footnote{In both the MAB algorithms and the outcome-generating models, feature values are represented using dummy variables.} 

\subsubsection{Outcome-generating models}

The outcome-generating model describes the \textit{true} relationship between student characteristics (feature values), the actions of assigning students to different versions of a technology, and the outcomes of student learning. We focus on scenarios in which two actions, such as choosing between concrete versus abstract explanations, affect the outcomes for two groups of students, such as those with math versus CS prerequisite as aforementioned. 

In each of the models, we generate the true reward probability for a student with particular features using logistic regression, with a separate logistic regression equation for each action.
Given a feature vector $x^{(j)}$ for student $j$, the reward probabilities are generated according to:
\begin{align*}
  P_{action=k}(\text{reward} = 1 \mid x^{(j)}) = \text{sigmoid}(b_{0,k} + \sum_{i=1}^{n}b_{i,k}x_{i}),
\end{align*}
where $b_k$ is the coefficient vector for action $k$ and has intercept $b_{0,k}$.
For our simulations, the coefficients for the feature values were zero for any feature past the first feature, meaning that a maximum of one student feature impacts the outcomes but more features may still be observed. By varying the coefficients for the intercept ($b_{0,k}$) and the first feature, we produced three models for the relationship among student characteristics (i.e., features or feature values), action choices, and outcomes (see Table~\ref{table:rewardProb}):

\vspace{-.05in}
\begin{small}
\begin{itemize}
  \item \textit{Baseline}: Student features have no impact on outcomes.
  \item \textit{Universal optimal action}: Student features have an impact on outcomes, but not the optimal action---the best version of the technology is the same regardless of features.
  \item \textit{Personalized optimal action}: Student features impact outcomes, meaning that the optimal action differs based on features---some students are better off experiencing Version A of the technology while for others Version B.
\end{itemize}
\end{small}
\vspace{-.05in}

For the baseline model,the coefficients of the actions vary only for the intercept in order to control the effect of each action when student features are ignored. For the universal optimal action model, we included four variations to capture different educationally meaningful scenarios. For instance, universal optimal action (1) reflects a case in which differences in prior knowledge minimally interact with the impact of different versions of a technological intervention, while (2) reflects a student characteristic magnifying the effectiveness of an intervention. 

\begin{table}[t]

\begin{tabular*}{\columnwidth}{@{\extracolsep{\fill}}lllll}
\toprule
\multicolumn{1}{r}{\textit{Relevant Feature:}}   & \multicolumn{2}{c}{F=0}         & \multicolumn{2}{c}{F=1}      \\
\multicolumn{1}{r}{\textit{Action Number:}}         & \multicolumn{1}{c}{A1} & \multicolumn{1}{c}{A2} & \multicolumn{1}{c}{A1} & \multicolumn{1}{c}{A2} \\
\midrule
Baseline     & 0.4     & \textbf{0.6}     & 0.4     & \textbf{0.6}     \\
Universal optimal action (1) & 0.4     & \textbf{0.6}     & 0.6     & \textbf{0.8}    \\
Universal optimal action (2) & 0.4     & \textbf{0.6}     & 0.4     & \textbf{0.8}\\
Universal optimal action (3) & 0.4     & \textbf{0.6}     & 0.5     & \textbf{0.7}\\
Universal optimal action (4) & 0.4     & \textbf{0.6}     & 0.8     & \textbf{0.9}     \\
Personalized optimal action  & 0.4     & \textbf{0.6}     & \textbf{0.6}     & 0.4    \\
\bottomrule
\end{tabular*}
\caption{Reward probabilities for each combination of actions (A1 and A2) and values of the relevant feature (F=0 and F=1) in the simulations. The optimal action, shown in bold, is the same (A2) for both feature values, except for the personalized optimal action model.}\label{table:rewardProb}
\end{table}

\subsubsection{Simulation parameters}

We varied three factors across the simulations: the outcome-generating model; the MAB algorithm (contextual or non-contextual);
and the number of student features. For all simulations, we considered three horizons: classrooms of 50, 250, and 1000 students. Multiple horizons illustrate the behavior of the algorithm at different time points and can guide decisions for incorporating adaptive algorithms based on the number of students who are expected to interact with the system. Each simulation was repeated 1000 times.

For the non-contextual Thompson sampling, parameters for a Beta distribution per action are learned independent of student features. For the contextual algorithm, we specify the weights of the student features as model coefficients. All simulations included at least one student feature regardless of the outcome-generating models.

To model the fact that curriculum designers may not know which student characteristics really matter, we included simulations where the observed features were a superset of those that actually impacted outcomes. Specifically, we considered models with a total of~1, 2, 3, 5, 7, 8, and 10 features. Therefore, for the non-baseline scenarios, the proportion of included features that impacted outcomes varied from 100\% to only 10\%. Since our contextual features are binary, we include indicator variables for each of the two values, and learn a separate weight for each indicator variable.\footnote{In pilot simulations, this encoding led to better performance than if only a single coefficient was learned for each feature, and corrected asymmetries in performance for students who had different values of the feature.}

\subsection{Results}\label{section:basicSimResults}

\begin{figure*}[ht]
    \centering
    \includegraphics[width=\textwidth]{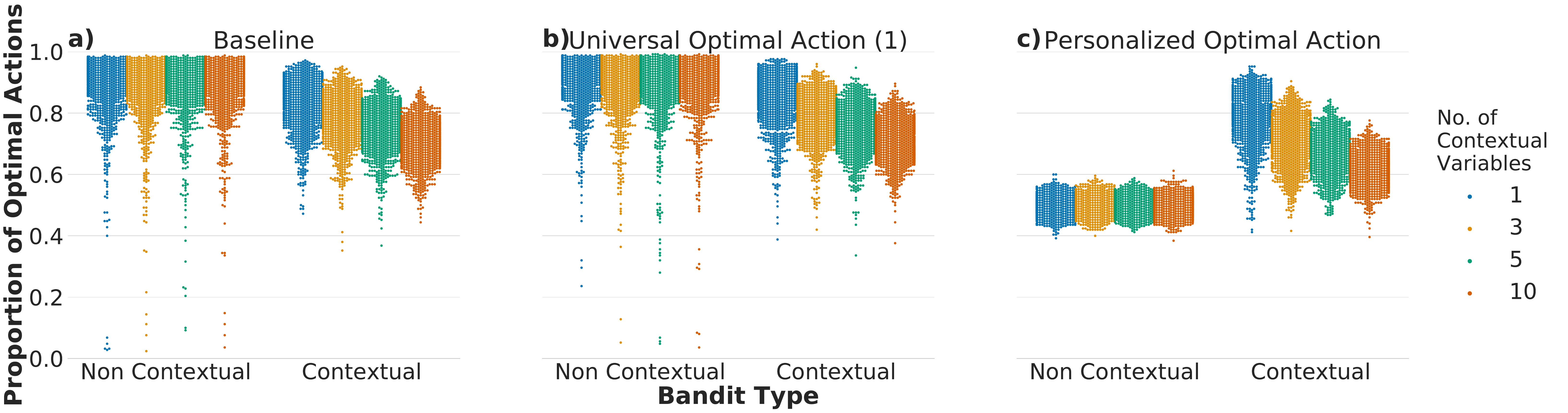}
    \caption{Swarm plots for the proportion of optimal actions for the two bandit types. Each point represents results from one trial with 250 students. For the universal optimal action, all scenarios show similar results; hence only scenario (1) is shown. The decreased performance of the contextual bandits in the baseline and universal optimal action scenarios, especially for large number of contextual variables, highlights the potential risks of personalization.}
    \label{fig:NumConVars}
\end{figure*}

\begin{table*}[ht]
\begin{tabular*}{\textwidth}{@{\extracolsep{\fill}}llrlrlr}
\toprule
{} & Superior bandit &  $|b|$ &         95\% CI &  $F(1, 13996)$ &       $p$ &  Cohen's $d$ \\
\midrule
Baseline                     &  Non Contextual &  0.058 &  [0.052, 0.064] &       6308.000 &  $< .001$ &        1.279 \\
Universal Optimal Action (1) &  Non Contextual &  0.054 &   [0.048, 0.06] &       6333.000 &  $< .001$ &        1.278 \\
Universal Optimal Action (2) &  Non Contextual &  0.051 &  [0.048, 0.054] &      15762.000 &  $< .001$ &        1.892 \\
Universal Optimal Action (3) &  Non Contextual &  0.053 &   [0.047, 0.06] &       5918.000 &  $< .001$ &        1.240 \\
Universal Optimal Action (4) &  Non Contextual &  0.057 &  [0.049, 0.065] &       3180.000 &  $< .001$ &        0.927 \\
Personalized Optimal Action  &      Contextual &  0.272 &  [0.269, 0.276] &      34180.000 &  $< .001$ &        2.610 \\
\bottomrule
\end{tabular*}

\caption{Inferential statistics for proportion of optimal actions for the two bandit types across all outcome-generating models, simulated for 1000 trials of 250 students each. $b$ represents the coefficient of improvement of results for the superior bandit after controlling for the number of contextual variables.}
\label{table:resultsNumConVars}

\end{table*}

\begin{figure}[ht]
    \centering
    \includegraphics[width=\columnwidth]{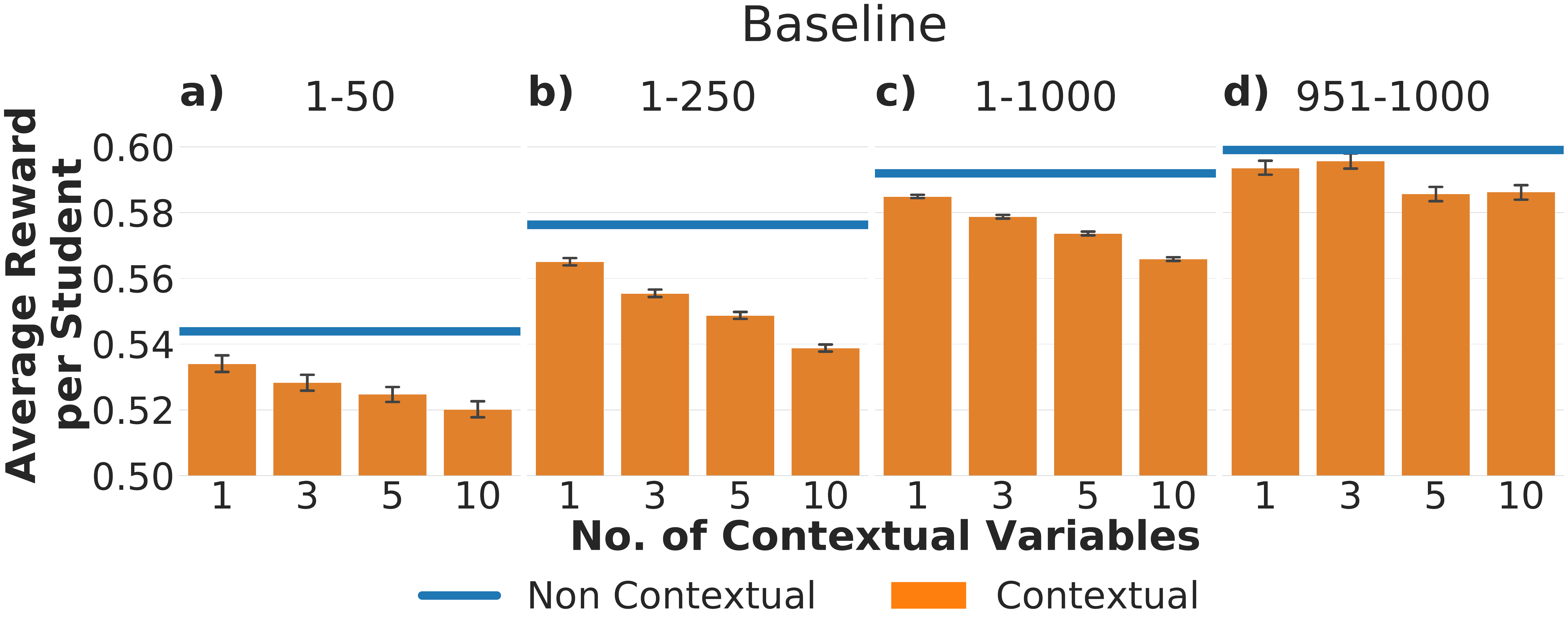}
    \caption{Average reward per student across 1--10 contextual variables for the two bandit types in the baseline model. In this model, the maximum possible expected reward is $0.6$, and the expected reward for uniform random assignment is $0.5$.
    Error bars represent 1 standard error.}
    \label{fig:NumConVarsRanges}
\end{figure}

First we focused on analyzing the performance of contextual and non-contextual MAB algorithms for the three outcome-generating models across 1 to 10 student features (i.e., contextual variables).
Using an analysis of covariance (ANCOVA), we compared the two MAB algorithms' performance with respect to the proportion of optimal actions for 250 students across 1000 trials, treating the number of contextual variables as a covariate.


\textbf{Baseline:} When student features do not influence outcomes, we see that as expected, the non-contextual bandit outperforms the contextual bandit (Table~\ref{table:resultsNumConVars}): average performance per student for the final 50 out of 1000 students using the contextual algorithm is similar to that of all 1000 students using the non-contextual algorithm (Figure~\ref{fig:NumConVarsRanges}). 
As the number of student features increases, the contextual MAB chooses a lower proportion of optimal actions for the first 250 students (Figure~\ref{fig:NumConVars}a), but the effect is relatively small especially when considering the impact on actual reward ($t(13996)=-37.654$, $p<0.001$, $b=-0.014$, 95\% CI = $[-0.014, -0.013]$). At longer horizons, the number of student features has less of an impact on overall average reward (Figure~\ref{fig:NumConVarsRanges}), which we discuss more below.

\textbf{Universal optimal action:} When outcomes are dependent on student features, the contextual MAB algorithm can learn a more accurate model than the non-contextual algorithm. However, when this more accurate model is not \textit{needed} for optimal action choices, learning the more accurate model does not improve action choices: the non-contextual bandit outperforms the contextual bandits in all four scenarios (Table~\ref{table:resultsNumConVars}; see Figure~\ref{fig:NumConVars}b for scenario 1).  While each scenario might arise due to different educational conditions, they are all very similar in how they appear to the non-contextual bandit algorithm.
The non-contextual bandit sees the two groups of students as identical, leading the overall performance to be the average for each group.

These changes in the average effectiveness of each intervention impact the algorithm's performance but do not necessarily degrade that performance; instead, the impact is dependent on how similar the two interventions are in their expected outcomes and how close those expected outcomes are to $0.5$, where there is the most variance.

\textbf{Personalized optimal action:} When the best policy for individual students depends on their features, the contextual bandit significantly outperforms the non-contextual bandit (Table~\ref{table:resultsNumConVars}). When only one student feature is included, the contextual MAB algorithm chooses the optimal action 61\% of the time for the first 50 students; this increases to 88\% for the final 50 of the total 250. Including extra student features decreases performance - if ten features are included and only one impacts the policy, the overall proportion of optimal actions falls to only 53\% for the first 50 students and 68\% for the final 50 students. Yet, this is still an improvement over the non-contextual algorithm (Figure~\ref{fig:NumConVars}c), which cannot exceed 50\% optimal actions in this scenario.
These results suggest that even if the number of students who will interact with the system is not large and one is uncertain about which of a limited set of features will impact the result, including those features will on average have a positive impact on student outcomes if one is confident that the best version of the system for an individual student varies based on one of those features. However, the size of the student population matters in considering this tradeoff: with the ``small'' population of 50 students, the most expressive contextual model was barely above chance performance. Each additional student characteristic brings a cost, and if one is using an adaptive algorithm in a real educational technology, one must carefully consider the chance that any characteristic will actually influence which version of the technology is best, rather than including all characteristics that might possibly influence outcomes.

\textbf{Variability in policies across students}:
As noted above, the extra parameters learned by the contextual MAB algorithm lead to the potential for greater variability in action choices within a single simulation. This can systematically affect groups of students when the algorithm attaches spurious relevance to a feature that does not actually impact outcomes. We can see this pattern by examining differences in action probabilities for students who differ only by characteristics that do not impact outcomes: that is, considering all students who have the same value for the first feature, how does the probability of choosing a particular action change based on their different values for the other features?
As the number of contextual variables increases, the average maximum difference in action choice probability between such students also increases from 9.8--15.6\% when two student features are included in the model to over 63.7--86.6\% when ten features are included in the model after running through 250 students. This occurs both based on the greater expressivity of the model with more student features and the fact that the model with more student features is likely still learning about the impact of each of these features.
This raises potential concerns about inequity: students who should be treated identically by the system may instead be treated systematically differently, based on features that do not impact how they learn. With both two and ten student characteristics, the largest variability was in the personalized optimal action scenario, suggesting that the benefits of the extra expressivity were not likely to be achieved for all students.

\section{Impact of Uneven Distribution\\ of Student Characteristics} 

\begin{figure}[t]
    \centering
    \includegraphics[width=\columnwidth]{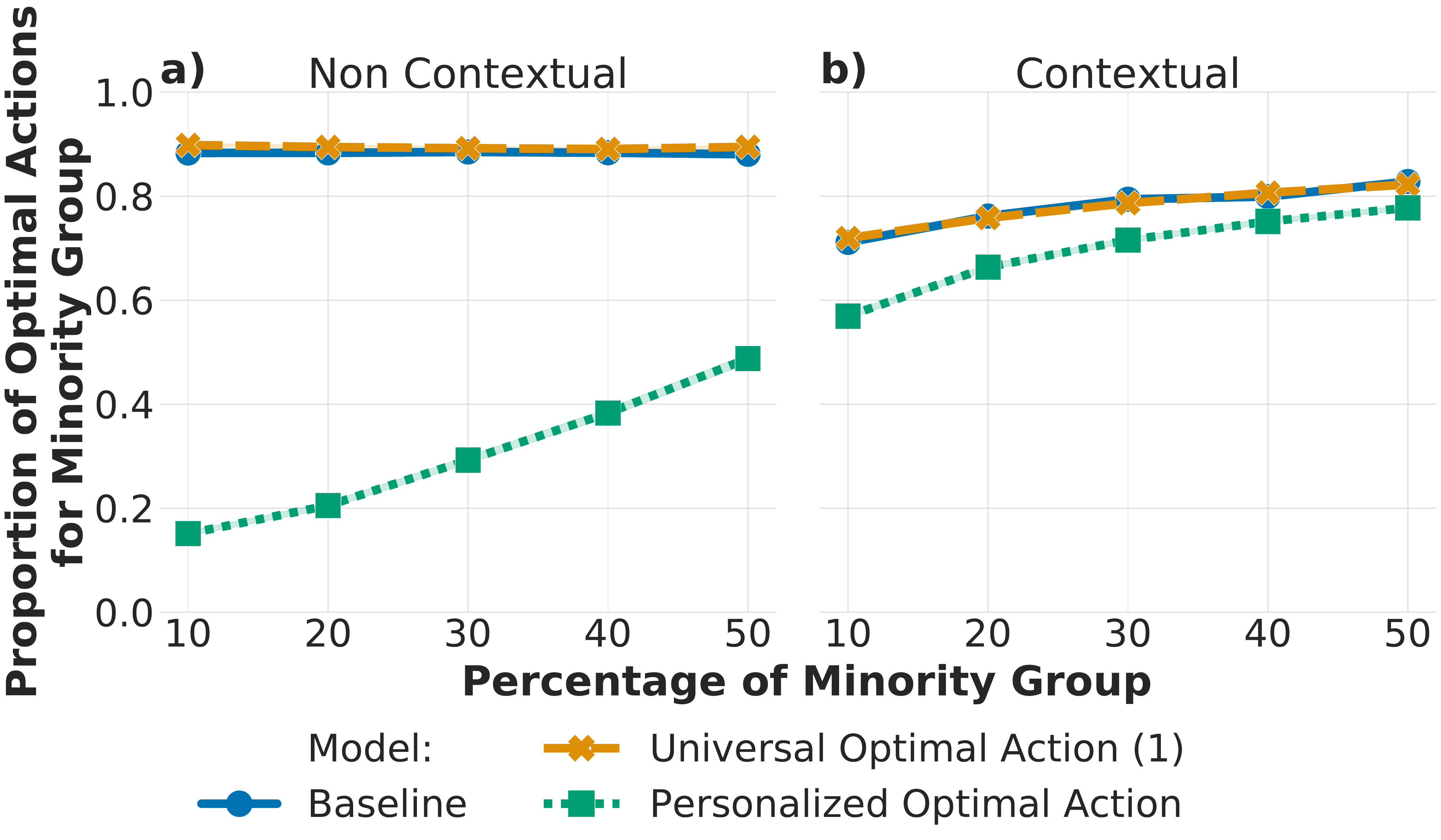}
    \caption{Proportion of optimal actions for minority groups with sizes of 10\%--50\% for the two bandit types across the three outcome-generating models, limited to one contextual variable. Standard errors, represented by the translucent bands, are negligible.}
    \label{fig:MinGrpSize1ConVar}
\end{figure}

The results of the previous simulations demonstrate that in situations where student characteristics (features) impact the outcome of different educational interventions, a contextual MAB algorithm only provides an improvement over a non-contextual algorithm when knowledge about the characteristic is necessary for choosing the best action. These simulations provided insight into how performance is impacted by different patterns of relationships between student characteristics and outcomes, with the assumption that those characteristics were uniformly distributed. However, in reality, some characteristics are likely to be more common than others. For example, when optimizing which hint to give to students who answer a question incorrectly, the algorithm is more likely to encounter a student with lower prior knowledge than one with higher prior knowledge.
Thus we now relax this assumption and explore how changing the distribution of student characteristics impacts student outcomes for both types of MAB algorithms. In these simulations, we examined not only overall outcomes, but also outcomes for different groups of students. Attention to group-specific outcomes is vital for identifying inequitable impacts of adaptive algorithms. 

\subsection{Methods}

Similar to the first set of simulations, we compared non-contextual and contextual MAB implementations that used Thompson sampling across the same three horizons of 50, 250, and 1000 students, with a focus on 250; we repeated each simulation 1000 times. These simulations include a new independent variable: the proportion of students in each group. Specifically, for each simulated student, we varied the probability of the student being in the minority group (i.e., having a value of one for the first student characteristic) from 10\% to 50\%, using 10\% increments. 
In addition to analyzing performance across all students, we examined performance for both the minority and majority groups separately. We also examined the \textit{balanced success rate}, defined as the simple average of the group-specific performances~\cite{ben2010user}. Balanced success rate provides a way of examining performance that treats each group as equally important, even though one group may have more students than another.

\begin{figure*}[t]
    \centering
    \includegraphics[width=\textwidth]{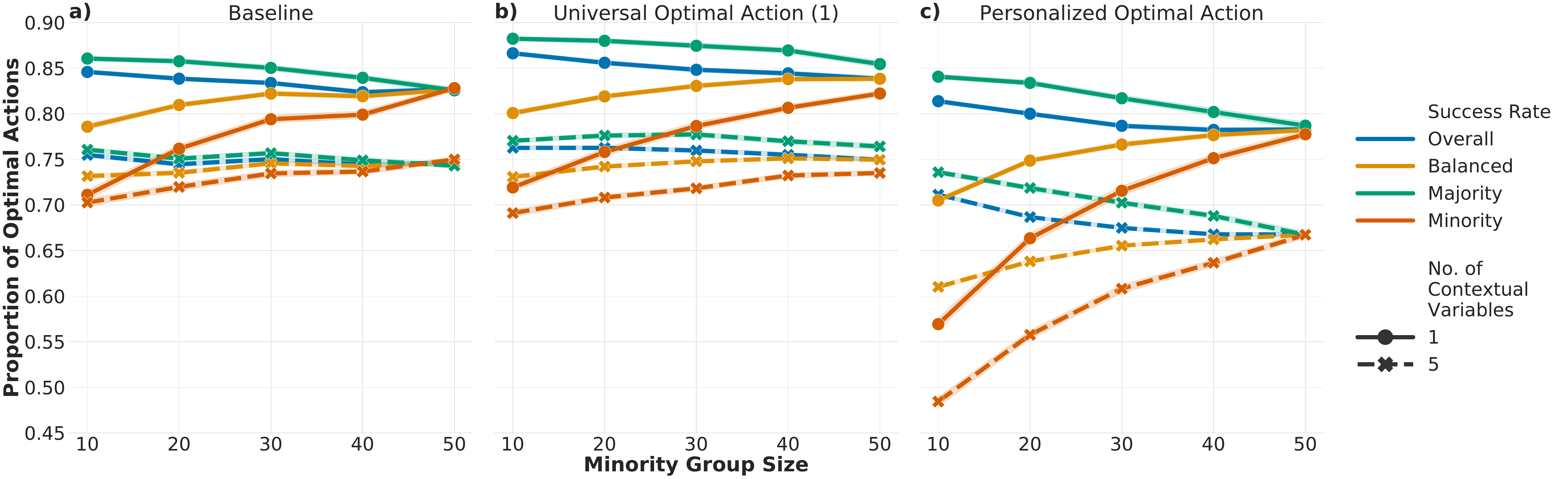}
    \caption{Comparing the proportion of optimal actions of the contextual bandit between 1 and 5 student features (i.e., contextual variables) for the majority and minority groups, as well as their balanced and overall averages, across minority group sizes of 10\%--50\%. Standard errors, represented by the translucent bands, are negligible.}
    \label{fig:Balanced_Success_Graph}
\end{figure*}
 
\subsection{Results}


As in the previous analysis, we used an ANCOVA to compare the performance for the two bandit types in terms of the proportion of optimal actions, but this time treating the percentage of the minority group as a covariate. 


\textbf{One student characteristic:} With one student characteristic, the contextual MAB algorithm's performance for the minority group decreases as the size of the minority group becomes smaller, across all outcome-generating models (Figure~\ref{fig:MinGrpSize1ConVar}b and Figure~\ref{fig:Balanced_Success_Graph}; $t(29988)=-18.894$, $p<0.001$, $b=-0.271$, 95\% CI = $[-0.300, -0.243]$).
This leads the contextual MAB algorithm to have a lower balanced success rate for smaller minority groups. However, overall performance across all students is slightly better since so many more students are in the majority group (Figure~\ref{fig:Balanced_Success_Graph}; $t(29988)=5.733$, $p<0.001$, $b=0.053$, 95\% CI = $[0.035, 0.071]$). In other words, decreasing the minority group size hurts the minority group more than it helps the majority group on a per-student basis; but replacing students from the minority group, who are assigned worse conditions, with students from the majority group, who are assigned better conditions, increases overall reward.  

This pattern of results occurs because the contextual MAB has more uncertainty about the impact of the particular value of the student characteristic that appeared fewer times: in the least balanced case, we expect the minority group to be seen only 25 times on average given a horizon of 250 students. Hence, providing a model with the potential to personalize for a minority group is a calculated risk - although the extra expressivity is likely intended to improve experiences for all groups of students, it can negatively impact minority groups, with a larger negative impact for smaller minority groups.

In contrast, the non-contextual MAB algorithm is relatively unaffected by the changing distribution of student characteristics in both the baseline ($t(9996)=-1.117$, $p=0.264$) and universal optimal action scenarios ($t(39996)=-0.358$, $p=0.721$), as shown by Figure~\ref{fig:MinGrpSize1ConVar}a. The changing distribution of student characteristics changes the expected rate of obtained reward from each action, but the changes are small enough that they have little impact on the algorithm's ability to choose optimal actions.

However, for the personalized optimal action model, the size of the minority group \textit{does} have a large impact on individual student outcomes for the non-contextual MAB algorithm: when the minority group is small, the algorithm learns to choose the action that is best for the majority and worst for the minority, resulting in the optimal action being chosen only 15\% of the time for the minority group, within a horizon of 250 students (Figure~\ref{fig:MinGrpSize1ConVar}a). When the two groups are of equal size, the algorithm has no systematic information that shows one action as consistently better or worse than the other; thus on average, it chooses the optimal action about 50\% of the time for both groups.

\textbf{Additional student characteristics for the contextual MAB algorithm:} When the number of student characteristics increases, the impact in the baseline and universal optimal actions models is relatively similar regardless of the size of the minority group. Balanced success rate is decreases by a relatively small amount with the addition of more student characteristics, ranging from an average decrease of 5.5 to 8.3 percentage points. 

For the personalized optimal action scenario, increasing the number of student characteristics from one to five also decreases performance, but does so more acutely: the average decrease in balanced success rate is 11 percentage points for this scenario. The impact for small minority groups is particularly great: with five student characteristics and a minority group size of 10\%, the optimal action is chosen just under half the time for the minority group, while with one student characteristic, the rate improved by 8.5 percentage points to 57\%. Differences in optimal action proportions were about half of this size for the other scenarios. This illustrates the differences in what must be learned: for the personalized optimal action scenario, increasing the number of characteristics makes the challenge of identifying which characteristic(s) matter even more challenging, especially with limited data.

\section{Real-World Experiments}

\begin{table*}[t]
\label{table:realworldSimSetups}
\centering
\begin{tabular*}{\textwidth}{lr@{\extracolsep{\fill}}rllll}
    \toprule
                                & Problem Set & Students & Q1 Size      & Q2 Size      & Q3 Size     & Q4 Size     \\
    \midrule
    Uneven Student Distribution & 293151      & 320      & 113 (35.3\%) & 100 (31.3\%) & 69 (21.6\%) & 38 (11.9\%) \\
    Even Student Distribution   & 263057      & 129      & 33 (25.6\%)  & 28  (21.7\%) & 34 (26.4\%) & 34 (26.4\%) \\
    \bottomrule
\end{tabular*}
\caption{Student totals and group distributions in the original ASSISTments data \cite{selent2016assistments} for the two problem sets of interest. Prior percent correct is discretized before removing students who have never answered incorrectly to experience the assigned condition, biasing group size towards the lower quartiles in the Uneven Student Distribution.}
\end{table*}

\begin{figure*}[t]
    \centering
    \includegraphics[width=\textwidth]{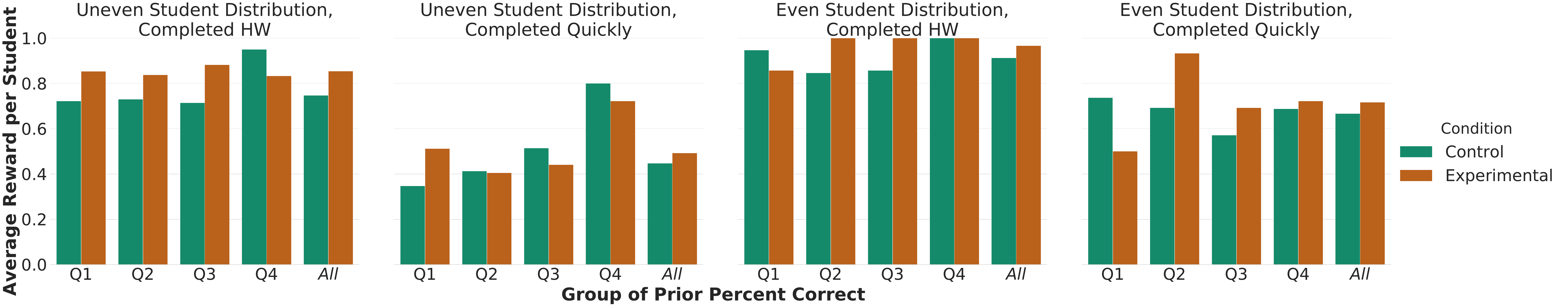}
    \caption{Original average reward per student in the ASSISTments data \cite{selent2016assistments}, across the four quartiles (Q1--Q4) of prior percent correct and their averages, for the two conditions in the experiments (control and experimental) illustrates our model parameters of real-world scenarios.}
    \label{fig:realworldParameters}
\end{figure*}

The first two sets of simulations can guide system designers when making decisions about personalizing based on student features. However, they have some limitations: while they considered a relatively large space of possibilities for how outcomes relate to student features, they focused on showing a general variety of cases rather than on specific cases that might be most common or of particular interest in education. To address this, we conducted several case studies of how MAB algorithms would have impacted actual experiments. We consider existing experimental data in which the optimal action would be personalized to see if the contextual MAB algorithms benefits students (as would be expected from our previous simulations) and also to demonstrate how factors from the previous simulations manifest in real-world scenarios. 

The experiments were previously conducted within \textit{ASSISTments}, an online learning system, and focused primarily on middle school math. We selected several experiments from~\cite{selent2016assistments}  based on how student outcomes were related to their prior successes in the system as well as their assignment of either the control or experimental condition. Prior success in the system is a strong candidate to be a student feature for personalization: it is typically easily available and can serve as a proxy for prior knowledge, which has been shown to influence the success of different instructional strategies~\cite{shute2008focus}. 

\subsection{Methods}

To model previously collected ASSISTments data in our MAB framework, we (1) transformed both the student characteristics and the student outcomes into discrete variables,\footnote{MAB algorithms can handle non-categorical data, but we focus on the categorical case to mirror our prior simulations.} and (2) resampled from the data to generate outcomes when the MAB algorithm assigned a condition.

For step (1), we first discretized students' prior percent correct on problems within ASSISTments, the sole student feature that we included for personalization, into four quartiles: the 25\% of students who began the homework assignment with the lowest prior percent correct (Q1), then those in the 26--50\% range (Q2), and so on. The dataset contains some students who began the homework but were not assigned to a condition. Since the experiments in~\cite{selent2016assistments} mainly manipulated students' experiences (e.g. type of hint) when they answered a question incorrectly, students who have never answered incorrectly are not included in the experiment results (nor will the MAB algorithm make choices for them). However, they are included in the quartile cutoffs, which means that in the population of students with whom the MAB algorithm interacts, the number of students in each quartile may not be uniform.

We also chose and discretized the student outcome measures. These experiments included two different measures of student outcomes: whether each student completed the homework and the number of problems that each student answered in the homework. 
All experiments took place in the \textit{SkillBuilder} interface, where students must answer three consecutive problems correctly to complete the homework. 
Completion of homework (denoted \textit{Completed HW}) is already discrete and could easily be collected in real time; two of our simulations use this measure. 
However, it is relatively coarse, as the vast majority of students completed the homework.
Thus, we also used a discretized version of the number of problems to completion (denoted \textit{Completed Quickly}). If a student completes the homework, doing so in fewer problems is a better outcome than doing so in more problems. Outcomes were based on the median problem count for students who completed the homework. Students who completed the homework in the median number of problems or fewer had positive reward, while those who did not complete the homework or completed it more slowly had no reward.
Though for practical use prior data would be needed to select an appropriate cut point, using a cut point based on collected data in our simulations measures the performance of students more closely.

\begin{figure*}[t]
    \centering
    \includegraphics[width=\textwidth]{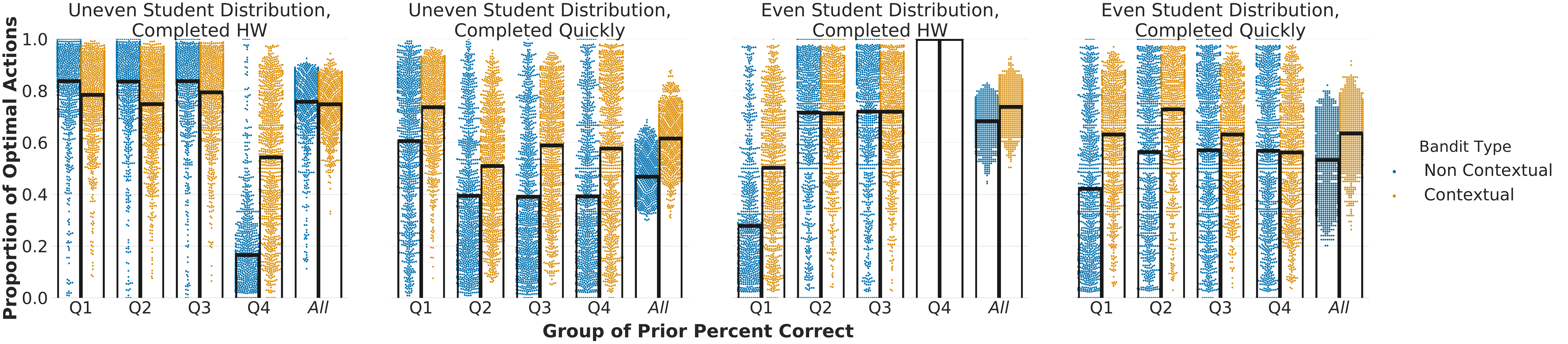}
    \caption{Swarm plots for the proportion of optimal actions for the two bandit types for each quartile of prior percent correct and their averages. Each point represents results from each of the 1000 trials per experiment and the solid black lines indicate the means of each swarm plots. Points for Q4 of Even Student Distribution, Completed HW are clustered at 1.0 because both actions are optimal. The extra information learned by the contextual bandit improves performance in most cases but the bimodality for some quartiles demonstrates the associated systematic risks.}
    \label{fig:realworldDistribution}
\end{figure*}

For step (2), we simulate a MAB algorithm's performance by repeatedly sampling students from the experiment. Within each trial, we fix the number of timesteps to the total number of students in the original experiment. At each timestep, a random student is sampled, and the algorithm then selects a condition for that student.
To compute the outcome, we sample from all outcomes for students in the experiment who were in the same quartile for prior percent correct and who experienced the same chosen condition. Each trial thus represents an experiment of the same size as the original, with the students drawn with replacement from the experimental data. We randomized each of the 1000 trials, though for each trial, we use the same student ordering for both types of MAB algorithms.

In our case studies, we focus on one problem set (\#293151) where students are unevenly distributed across quartiles, with more lower-performing students (Q1), and one problem set (\#263057) in which students are more evenly distributed across quartiles (see Table~\ref{table:realworldSimSetups}).  
With the two different outcome measures, this resulted in four simulation scenarios.
We chose these problem sets because they had student outcomes that varied based on both condition and student quartile (see Figure~\ref{fig:realworldParameters}), representing a setting where there is the most potential benefit from a contextual MAB algorithm.

\subsection{Results}\label{section:assistmentsResults}

In all four settings, at least one quartile of students (out of Q1--Q4) was helped by the contextual MAB algorithm, and in three of the four settings, average outcomes across all students were improved by personalization.

\textbf{Uneven Student Distribution, Completed HW:} As shown in Figure~\ref{fig:realworldDistribution}a, in this scenario, students in Q4 were much more likely to experience their optimal condition with a contextual MAB algorithm. This occurs because the condition that is best for the average student is the one that is worse for Q4: the non-contextual MAB thus optimizes in a way that has a systematic, negative outcome for Q4 students. Conversely, the contextual MAB algorithm does not do as well as the non-contextual algorithm for students in Q1--Q3 because of the extra exploration needed to learn about more variables that are not necessary to help these students. Overall, this means that the contextual MAB algorithm had a slightly lower rate of choosing the optimal action than the non-contextual MAB. However, the difference is only 1\% of students, and this difference is even smaller in terms of average reward: average reward is reduced by less than $0.005$ overall, while it is increased for Q4 students by about $0.044$. In this experiment, reward rates are high in general (greater than $70\%$ for all conditions and quartiles). Thus with 320 students, small differences in condition assignment often are not reflected in large differences in outcomes. Q1--Q3 students have very similar outcomes across the two methods of condition assignment; Q4 has the greatest difference in success for one condition versus another, and thus the large increase in optimal condition assignment for these students does boost their average outcomes. 

\textbf{Uneven Student Distribution, Completed Quickly:} Using the Completed Quickly outcome measure with the same students, students in all quartiles were more likely to be assigned to the optimal condition when the contextual MAB algorithm was used (Figure~\ref{fig:realworldDistribution}b). This pattern occurs because the overall probability of a positive outcome is very similar across the two conditions when student quartiles are ignored (shown by \textit{All} in Figure~\ref{fig:realworldParameters}b), making it difficult for the non-contextual bandit to learn that the experimental condition is better on average. In contrast, the differences between conditions are large for all quartiles except Q2. Thus, the information from the student quartiles makes the problem easier for the contextual MAB algorithm, though the relatively small difference between conditions for Q2 results in the lowest overall proportion of optimal action choices. This simulation thus importantly shows a scenario that was not explored in the prior simulations, in which knowing about extra information increases the number of  parameters to learn but makes learning about each of those individual parameters easier.

\textbf{Even Student Distribution, Completed HW:} For this scenario, there were again very high reward probabilities across all conditions, and a relatively small overall difference between conditions but larger differences between conditions for three of the four quartiles. Results here had some of the characteristics of each of the previous two simulations: Q1 students experienced the largest gain in optimal condition assignment from the contextual MAB algorithm, as the best condition for them was not the same as the best condition overall. Q2 and Q3 students had similar rates of optimal condition performance regardless of algorithm. With the non-contextual MAB algorithm, there were fewer parameters to learn about, but the difference between arms was smaller, making the problem harder; with the contextual MAB algorithm, the increase in parameters did increase the time to learn, but this negative impact was balanced by the fact that the difference between conditions was larger for Q2 and Q3 than for students overall. All Q4 students completed their homework, so no differences between algorithms were possible. Overall, the contextual MAB algorithm has slightly better performance than the non-contextual MAB algorithm due to the improvement for Q1 students.


\textbf{Even Student Distribution, Completed Quickly:} Finally, using the Completed Quickly outcome measure for this second set of students, the results were still largely in favor of the contextual MAB algorithm. As the experimental condition is better on average, students in Q1 experience a large positive impact through personalization because the control condition is uniquely better for them. Q2 and Q3 also experience positive impacts, with the impact on Q2 students being larger because the difference between the two conditions is larger, which speeds learning for the contextual bandit. Conversely, Q4 students experience slightly less positive outcomes under the contextual MAB algorithm because the small difference between conditions slows learning; in comparison, the non-contextual MAB algorithm is more beneficial for Q4 students since the overall difference between conditions across all students is larger than the difference for Q4 students only.

\section{Discussion}

Real-time adaptive algorithms can respond quickly to optimize experiences for individual students, and their expressivity for personalizing experiences increases with each additional type of student information they are given. In this paper, we have shown that this expressivity may be worthwhile under two circumstances: when it is \textit{necessary}  for expressing the best policy to improve student outcomes, as shown by the simulation results in Section~\ref{section:basicSimResults}, and in certain cases when including those characteristics magnifies differences between conditions for subgroups of students, as shown by the results in Section~\ref{section:assistmentsResults}. For the scenarios we considered, in which there are only two interventions and binary rewards, this magnification in differences can only occur on average if the optimal policy varies across students (or if there is a ceiling or floor effect for rewards for some subgroup of students), but it is possible this situation could occur in other circumstances with more complex environments.

Our results also show the potential ethical concerns inherent in choosing whether to include student characteristics when these characteristics are not uniformly distributed. If these characteristics matter, then failing to include them may lead to a learned policy that systematically optimizes for the majority but not for a minority group. However, when this expressivity is not necessary, it increases variability across students and also increases the time for identifying the correct policy, thus significantly decreasing the number of students assigned the best version of the technology and slightly decreasing their average outcomes. In these cases, a minority group may disproportionately bear the cost of the need for the algorithm to learn additional parameters.


There are several limitations to our results. First, we have focused only on discrete student features and discrete outcomes but continuous parameters are also common. For example, we might measure student scores rather than homework completion or model prior knowledge as an estimated ability parameter. If one wanted to extend these analyses to real-valued student features, one could easily incorporate them into the current modeling framework with versions of Thompson sampling for real-valued outcomes~\cite{agrawal2013thompson}, and there exist metrics from a large literature for assessing whether students are treated fairly (e.g.,~\cite{binns2018fairness}). Using real-valued parameters is unlikely to significantly impact trends in results, except that defining student groups for analyzing equitable outcomes is more difficult. Our results from our universal optimal action scenarios show that, with binary rewards, knowledge of the student features is not beneficial if it is unnecessary for expressing the best policy. However, these results may not translate to the real-valued rewards case, where the latent student features will add to the variability in the distributions observed by the non-contextual bandit, and exploring these scenarios is an important step for future work. 
A second limitation is that our simulations comprise only a single student feature that influences the outcome, though in actual deployments multiple features may influence the best policy. Still, we believe that our results can guide system designers when thinking about such scenarios, especially in weighing the costs and benefits of including each possible variable.


The results from the real-world scenarios highlight the potential value of MAB algorithms for educational technologies. For almost all scenarios and groups, both types of MAB algorithms chose the optimal condition more often than if students had been assigned uniformly at random, and average rewards were in many cases very close to the optimal expected reward (i.e. if the optimal action had been chosen for all students). The absolute difference in rewards was relatively small between the two bandit types--at most $0.075$--and the contextual bandit achieved at worst 12\% less than the optimal expected reward for any student group. Yet the earlier simulations urge caution for incorporating student characteristics, due to (1) decreases in achieved outcomes when these characteristics are unnecessary and (2) the systematically different treatment of students based on irrelevant characteristics, as illustrated by large difference in condition assignment probability between vectors of student characteristics with identical outcome probabilities. Thus, system designers should weigh the risk of not personalizing when the best policy for the minority differs from the majority with these side effects of personalization and ultimately strive to only include variables that past evidence suggests differentially impact outcomes.

One could make a number of extensions of this work for using MAB algorithms to improve and personalize educational technologies. First, contextual MAB algorithms might mitigate issues of biases when different types of students interact with an educational technology and while all are most helped by the same version of the technology, their outcomes have different distributions. For example, struggling students may complete homework later, leading the MAB algorithm's early estimates to be non-representative of the broader population. Prior work has shown that this bias significantly worsens inference about the effectiveness of the technology as well as expected student outcomes~\cite{rafferty2019statistical}: the use of a contextual MAB algorithm could allow the system to adapt to such differences across students. Second, if the technology is used by a large number of students, the set of variables used by the contextual algorithm could be increased as more data are collected. Such a system might improve consistency across student outcomes, while still personalizing based on truly relevant features that are justified the sufficient information collected. The work in this paper both provides a starting point for considering what scenarios, algorithms, and metrics should be explored in future work, as well as guidance for system designers who would like to deploy MAB algorithms within their own technologies but are uncertain about which student characteristics, if any, to include for personalization. While including all possible characteristics might capture the desire for maximum peronalization, this works points to the potential costs of such personalization and suggests that consideration of how likely it is that such characteristics will matter is important both for performance and equitability across students.



\bibliographystyle{abbrv}
\bibliography{CompiledBib}

\begin{thebibliography}{10}

\bibitem{agrawal2012Analysis}
S.~Agrawal and N.~Goyal.
\newblock Analysis of {T}hompson sampling for the multi-armed bandit problem.
\newblock In S.~Mannor, N.~Srebro, and R.~C. Williamson, editors, {\em
  Proceedings of the 25th Annual Conference on Learning Theory}, volume~23,
  pages 39.1--39.26, Edinburgh, Scotland, 2012. PMLR.

\bibitem{agrawal2013thompson}
S.~Agrawal and N.~Goyal.
\newblock Thompson sampling for contextual bandits with linear payoffs.
\newblock In S.~Dasgupta and D.~McAllester, editors, {\em Proceedings of the
  30th International Conference on International Conference on Machine
  Learning}, volume~28, pages 127--135. JMLR, 2013.

\bibitem{aleven2016help}
V.~Aleven, I.~Roll, B.~M. McLaren, and K.~R. Koedinger.
\newblock Help helps, but only so much: Research on help seeking with
  intelligent tutoring systems.
\newblock {\em International Journal of Artificial Intelligence in Education},
  26(1):205--223, 2016.

\bibitem{auer2002using}
P.~Auer.
\newblock Using confidence bounds for exploitation-exploration trade-offs.
\newblock {\em Journal of Machine Learning Research}, 3(Nov):397--422, 2002.

\bibitem{ben2010user}
A.~Ben-Hur and J.~Weston.
\newblock A user's guide to support vector machines.
\newblock In {\em Data mining techniques for the life sciences}, pages
  223--239. Springer, 2010.

\bibitem{binns2018fairness}
R.~Binns.
\newblock Fairness in machine learning: Lessons from political philosophy.
\newblock In {\em Conference on Fairness, Accountability and Transparency},
  pages 149--159, 2018.

\bibitem{chapelle2011empirical}
O.~Chapelle and L.~Li.
\newblock An empirical evaluation of {T}hompson sampling.
\newblock In J.~Shawe-Taylor, R.~S. Zemel, P.~L. Bartlett, F.~Pereira, and
  K.~Q. Weinberger, editors, {\em Advances in Neural Information Processing
  Systems 24}, pages 2249--2257. Curran Associates, Inc., 2011.

\bibitem{clement2015multi}
B.~Clement, D.~Roy, P.-Y. Oudeyer, and M.~Lopes.
\newblock Multi-armed bandits for intelligent tutoring systems.
\newblock {\em {Journal of Educational Data Mining}}, 7:20--48, 2015.

\bibitem{corbett2001cognitive}
A.~T. Corbett.
\newblock Cognitive computer tutors: {S}olving the two-sigma problem.
\newblock In M.~Bauer, P.~Gmytrasiewicz, and J.~Vassileva, editors, {\em User
  Modeling 2001}, volume 2109 of {\em Lecture Notes in Computer Science}, pages
  137--147. Springer, 2001.

\bibitem{gardner2019evaluating}
J.~Gardner, C.~Brooks, and R.~Baker.
\newblock Evaluating the fairness of predictive student models through slicing
  analysis.
\newblock In {\em Proceedings of the 9th International Conference on Learning
  Analytics \& Knowledge}, pages 225--234, 2019.

\bibitem{hajian2016algorithmic}
S.~Hajian, F.~Bonchi, and C.~Castillo.
\newblock Algorithmic bias: From discrimination discovery to fairness-aware
  data mining.
\newblock In {\em Proceedings of the 22nd ACM SIGKDD international conference
  on knowledge discovery and data mining}, pages 2125--2126, 2016.

\bibitem{hutt2019evaluating}
S.~Hutt, M.~Gardner, A.~L. Duckworth, and S.~K. D'Mello.
\newblock Evaluating fairness and generalizability in models predicting on-time
  graduation from college applications.
\newblock {\em International Educational Data Mining Society}, 2019.

\bibitem{jain2014multiarmed}
S.~Jain, B.~Narayanaswamy, and Y.~Narahari.
\newblock A multiarmed bandit incentive mechanism for crowdsourcing demand
  response in smart grids.
\newblock In {\em Twenty-Eighth AAAI Conference on Artificial Intelligence},
  2014.

\bibitem{joseph2016fairness}
M.~Joseph, M.~Kearns, J.~H. Morgenstern, and A.~Roth.
\newblock Fairness in learning: Classic and contextual bandits.
\newblock In {\em Advances in Neural Information Processing Systems}, pages
  325--333, 2016.

\bibitem{lan2016contextual}
A.~S. Lan and R.~G. Baraniuk.
\newblock A contextual bandits framework for personalized learning action
  selection.
\newblock In T.~Barnes, M.~Chi, and M.~Feng, editors, {\em Proceedings of the
  Ninth International Conference on Educational Data Mining}, pages 424--429,
  2016.

\bibitem{li2010contextual}
L.~Li, W.~Chu, J.~Langford, and R.~E. Schapire.
\newblock A contextual-bandit approach to personalized news article
  recommendation.
\newblock In {\em Proceedings of the 19th International Conference on World
  Wide Web}, pages 661--670. ACM, 2010.

\bibitem{liu2014trading}
Y.-E. Liu, T.~Mandel, E.~Brunskill, and Z.~Popovic.
\newblock Trading off scientific knowledge and user learning with multi-armed
  bandits.
\newblock In J.~Stamper, Z.~Pardos, M.~Mavrikis, and B.~McLaren, editors, {\em
  Proceedings of the 7th International Conference on Educational Data Mining},
  pages 161--168, 2014.

\bibitem{lomas2016interface}
J.~D. Lomas, J.~Forlizzi, N.~Poonwala, N.~Patel, S.~Shodhan, K.~Patel,
  K.~Koedinger, and E.~Brunskill.
\newblock Interface design optimization as a multi-armed bandit problem.
\newblock In {\em Proceedings of the 2016 CHI conference on human factors in
  computing systems}, pages 4142--4153, 2016.

\bibitem{rafferty2019statistical}
A.~N. Rafferty, H.~Ying, and J.~Williams.
\newblock Statistical consequences of using multi-armed bandits to conduct
  adaptive educational experiments.
\newblock {\em Journal of Educational Data Mining}, 11(1):47--79, 2019.

\bibitem{pmlr-v75-raghavan18a}
M.~Raghavan, A.~Slivkins, J.~V. Wortman, and Z.~S. Wu.
\newblock The externalities of exploration and how data diversity helps
  exploitation.
\newblock In S.~Bubeck, V.~Perchet, and P.~Rigollet, editors, {\em Proceedings
  of the 31st Conference On Learning Theory}, volume~75 of {\em Proceedings of
  Machine Learning Research}, pages 1724--1738. PMLR, 06--09 Jul 2018.

\bibitem{segal2018combining}
A.~Segal, Y.~B. David, J.~J. Williams, K.~Gal, and Y.~Shalom.
\newblock Combining difficulty ranking with multi-armed bandits to sequence
  educational content.
\newblock In {\em Proceedings of the 19th International Conference on
  Artificial Intelligence in Education}, pages 317--321. Springer, 2018.

\bibitem{selent2016assistments}
D.~Selent, T.~Patikorn, and N.~Heffernan.
\newblock {ASSISTments} dataset from multiple randomized controlled
  experiments.
\newblock In {\em Proceedings of the Third {ACM} Conference on Learning at
  Scale}, pages 181--184. ACM, 2016.

\bibitem{shaikh2019balancing}
H.~Shaikh, A.~Modiri, J.~J. Williams, and A.~N. Rafferty.
\newblock {Balancing Student Success and Inferring Personalized Effects in
  Dynamic Experiments}.
\newblock In {\em {Proceedings of the 12th International Conference on
  Educational Data Mining}}, 2019.

\bibitem{shute2008focus}
V.~Shute.
\newblock Focus on formative feedback.
\newblock {\em {Review of Educational Research}}, 78(1):153--189, 2008.

\bibitem{villar2015multi}
S.~S. Villar, J.~Bowden, and J.~Wason.
\newblock Multi-armed bandit models for the optimal design of clinical trials:
  {B}enefits and challenges.
\newblock {\em Statistical science: {A} review journal of the Institute of
  Mathematical Statistics}, 30(2):199--215, 2015.

\bibitem{williams2016axis}
J.~J. Williams, J.~Kim, A.~N. Rafferty, S.~Maldonado, K.~Z. Gajos, W.~S.
  Lasecki, and N.~Heffernan.
\newblock Axis: Generating explanations at scale with learnersourcing and
  machine learning.
\newblock In {\em Proceedings of the Third {ACM} Conference on Learning at
  Scale}, pages 379--388. ACM, 2016.

\bibitem{williams2018enhancing}
J.~J. Williams, A.~N. Rafferty, D.~Tingley, A.~Ang, W.~S. Lasecki, and J.~Kim.
\newblock Enhancing online problems through instructor-centered tools for
  randomized experiments.
\newblock In {\em Proceedings of the 2018 CHI Conference on Human Factors in
  Computing Systems}, pages 207:1--207:12. ACM, 2018.

\end{thebibliography}

\balancecolumns
\end{document}